\pdfoutput=1

\documentclass[11pt]{article}

\usepackage{acl}
\usepackage{amsmath}
\usepackage{times}
\usepackage{paralist}
\usepackage{algorithm}
\usepackage{algpseudocode}
\usepackage{amsfonts}

\usepackage{multirow}
\usepackage{latexsym}
\usepackage{graphicx}
\usepackage{bm}
\usepackage{contour}
\usepackage{adjustbox}
\usepackage{mathrsfs}
\usepackage{booktabs,caption}
\usepackage[flushleft]{threeparttable}
\usepackage{fancyhdr}

\usepackage[T1]{fontenc}

\usepackage[utf8]{inputenc}

\usepackage{microtype}
\DeclareMathOperator*{\argmax}{argmax}
\usepackage[english]{babel}
\usepackage{amsthm}
\newtheorem{theorem}{Theorem}

\usepackage{changes}

\title{Learning Non-Autoregressive Models from Search\\ for Unsupervised Sentence Summarization}

\author{Puyuan Liu, Chenyang Huang, Lili Mou \\
Dept.~Computing Science, Alberta Machine Intelligence Institute (Amii)\\University of Alberta, Canada \\ \texttt{\{puyuan, chuang8\}@ualberta.ca, doublepower.mou@gmail.com}}

\begin{document}
\setlength{\abovedisplayskip}{7pt}
\setlength{\belowdisplayskip}{7pt}
\setlength{\abovedisplayshortskip}{7pt}
\setlength{\belowdisplayshortskip}{7pt}

	\setlength{\textfloatsep}{10pt plus 1.0pt minus 2.0pt}
\setlength{\floatsep}{10pt plus 1.0pt minus 2.0pt}
\setlength{\intextsep}{10pt plus 1.0pt minus 2.0pt}

\maketitle
\begin{abstract}

Text summarization aims to generate a short summary for an input text. 
In this work, we propose a Non-Autoregressive Unsupervised Summarization (NAUS) approach, which does not require parallel data for training. Our NAUS first performs edit-based search towards a heuristically defined score, and generates a summary as pseudo-groundtruth. Then, we train an encoder-only non-autoregressive Transformer based on the search result. We also propose a dynamic programming approach for length-control decoding, which is important for the summarization task. Experiments on two datasets show that NAUS achieves state-of-the-art performance for unsupervised summarization, yet largely improving inference efficiency. Further, our algorithm is able to perform explicit length-transfer summary generation.\footnote{Our code, model, and output are released at: \url{https://github.com/MANGA-UOFA/NAUS}}

\end{abstract}

\section{Introduction}
\renewcommand{\headrulewidth}{0pt}
\cfoot{In \textit{ACL}, pages 7916-7929, 2022}
\thispagestyle{fancy} 

Text summarization is an important natural language processing (NLP) task, aiming at generating concise summaries for given texts while preserving the key information. 
It has extensive real-world applications such as headline generation~\cite{nenkova2011automatic}. {In this paper, we focus on the setting of sentence summarization~\cite{Rush_2015,filippova-etal-2015-sentence}}.

State-of-the-art text summarization models are typically trained in a supervised way with large training corpora, comprising pairs of long texts and their summaries \cite{zhang2020pegasus, aghajanyan2020better, aghajanyan2021muppet}.
However, such parallel data are expensive to obtain, preventing the applications to less popular domains and less spoken languages.

Unsupervised text generation has been attracting increasing interest, because it does not require parallel data for training.  
One widely used approach is to compress a long text into a short one, and to reconstruct it to the long text by a cycle consistency loss \cite{miao2016language,wang-lee-2018-learning,baziotis-etal-2019-seq}.
Due to the indifferentiability of the compressed sentence space, such an approach requires reinforcement learning (or its variants), which makes the training difficult \cite{kreutzer-etal-2021-offline}.

Recently, \newcite{schumann-etal-2020-discrete} propose an edit-based approach for unsupervised summarization. Their model maximizes a heuristically defined scoring function that evaluates the quality (fluency and semantics) of the generated summary, achieving higher performance than cycle-consistency methods. 
However, the search approach is slow in inference because hundreds of search steps are needed for each data sample.
Moreover, their approach can only select words from the input sentence with the word order preserved. Thus, it is restricted and may generate noisy summaries due to the local optimality of search algorithms. 

To address the above drawbacks, we propose a Non-Autoregressive approach to Unsupervised Summarization (NAUS). 
The idea is to perform search as in \newcite{schumann-etal-2020-discrete} and, inspired by \newcite{NEURIPS2020_7a677bb4}, to train a machine learning model to smooth out such noise and to speed up the inference process.
Different from \newcite{NEURIPS2020_7a677bb4}, we propose to utilize \textit{non-autoregressive} decoders, which generate all output tokens in parallel due to our following observations:

$\bullet$ Non-autoregressive models are several times faster than autoregressive generation, which is important when the system is deployed.

$\bullet$ The input and output of the summarization task have a strong correspondence. Non-autoregressive generation supports encoder-only architectures, which can better utilize such input--output correspondence and even outperform autoregressive models for summarization.

$\bullet$ For non-autoregressive models, we can design a length-control algorithm based on dynamic programming to satisfy the constraint of output lengths, which is typical in summarization applications but cannot be easily achieved with autoregressive models.

We conducted experiments on Gigaword headline generation \cite{graff2003english} and DUC2004
\cite{duc2004} datasets. Experiments show that our NAUS achieves state-of-the-art performance on unsupervised summarization; 
especially, it outperforms its teacher (i.e., the search approach), confirming that NAUS can indeed smooth out the search noise. Regarding inference efficiency, our NAUS with truncating is 1000 times more efficient than the search approach; even with dynamic programming for length control, NAUS is still 100 times more efficient than search and several times more efficient than autoregressive models. Our NAUS is also able to perform length-transfer summary generation, i.e., generating summaries of different lengths from training.

\section{Approach}

\begin{figure*}[!t]
\centering
\includegraphics[width=\linewidth]{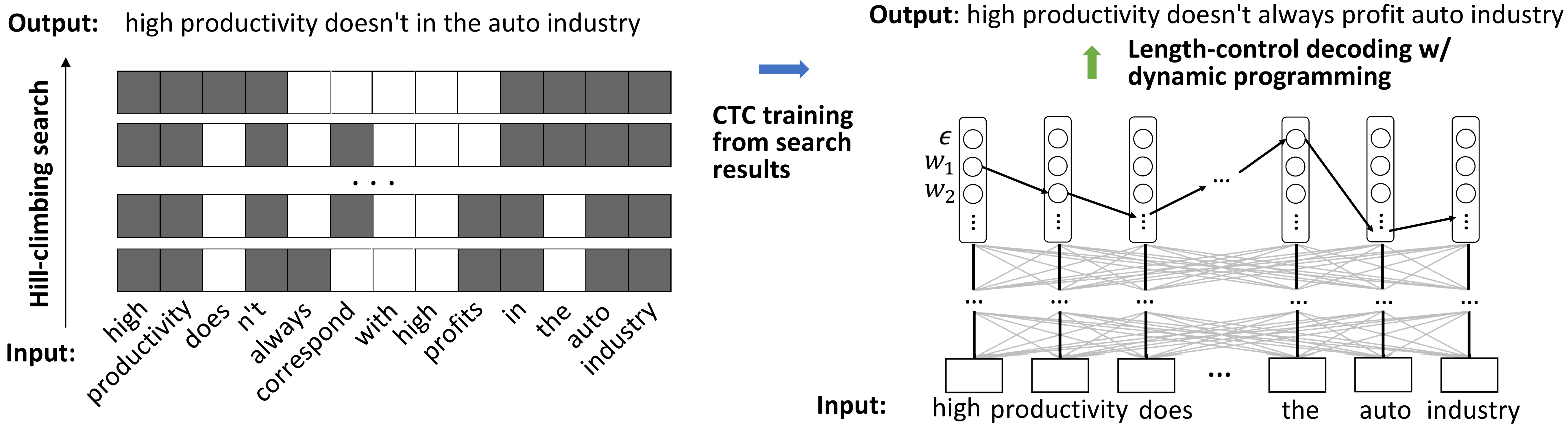}
\caption{The overview of our NAUS approach. In each search step, input words corresponding to grey cells are selected. 
Besides, the blue arrow refers to the training process, and the green arrow refers to inference.}
\end{figure*}

In our approach, we first follow \newcite{schumann-etal-2020-discrete} and obtain a summary by discrete search towards a heuristically defined objective function  (\S\ref{ss:search}). Then, we propose a non-autoregressive model for the summarization task (\S\ref{ss:NAG}). 
We present the training strategy and the proposed length-control algorithm in \S\ref{ss:DP}.

\subsection{Search-Based Summarization}\label{ss:search}

Consider a given source text $\mathbf{x}=\left(\mathrm x_{1},\mathrm  x_{2}, \ldots, \mathrm x_{n}\right)$. The goal of summarization is to find a shorter text $\mathbf{y}=\left(\mathrm y_{1}, \mathrm y_{2}, \ldots,\mathrm  y_{m}\right)$ as the summary.

Our work on unsupervised summarization follows the recent progress of search-based text generation~\cite{liu-etal-2020-unsupervised,liu2021simulated,kumar2020iterative}. \newcite{schumann-etal-2020-discrete} formulate summarization as word-level extraction (with order preserved), and apply edit-based discrete local search to maximize a heuristically designed objective.

Specifically, the objective function considers two aspects: 
(1) a language fluency score $f_{{\mathrm{LM}}}(\mathbf{y})$, given by the reciprocal of a language model's perplexity; and (2) a semantic similarity score $f_{\mathrm{SIM}}(\mathbf{y}; \mathbf{x})$, given by the cosine embeddings.
The overall objective combines the two aspects as
\begin{align}\label{eqn:score}
f(\mathbf{y} ; \mathbf{x})=f_{{\mathrm{LM}}}(\mathbf{y}) \cdot f_{\mathrm{SIM}}(\mathbf{y} ; \mathbf{x})^{\gamma}
\end{align}
where $\gamma$ is a weighting hyperparameter. Interested readers are referred to \newcite{schumann-etal-2020-discrete} for the details of the scoring function.

Further, the desired summary length can be specified as a hard constraint, achieved by searching only among sentences of the correct length. Suppose the desired summary length is $T$, the approach selects $T$ random words from the input, and maximizes the scoring function~\eqref{eqn:score} by changing the selection and non-selection of two words.

A greedy hill-climbing algorithm determines whether the change is accepted or not.
In other words, a change is accepted if the score improves, or rejected otherwise. Such a process continues until a (possibly local) optimum is found.

A pilot analysis in \newcite{schumann-etal-2020-discrete} shows that words largely overlap between a source text and its reference summary. 
This explains the high performance of such a word extraction approach, being a state-of-the-art unsupervised summarization system and outperforming strong competitors, e.g., cycle consistency \cite{wang-lee-2018-learning, baziotis-etal-2019-seq}. 

\subsection{Non-Autoregressive Model for Summarization}\label{ss:NAG}
Despite the high performance, such edit-based search has several drawbacks. 
First, the search process is slow because hundreds of local search steps are needed to obtain a high-quality summary. Second, their approach only extracts the original words with order preserved. Therefore, the generated summary is restricted and may be noisy.

To this end, we propose a Non-Autoregressive approach to Unsupervised Summarization (NAUS) by learning from the search results. 
In this way, the machine learning model can smooth out the search noise and is much faster, largely alleviating the drawbacks of search-based summarization. 
Compared with training an autoregressive model from search~\cite{NEURIPS2020_7a677bb4}, non-autoregressive generation predicts all the words in parallel, further improving inference efficiency by several times.

Moreover, a non-autoregressive model enables us to design an encoder-only architecture, which is more suited to the summarization task due to the strong correspondence between input and output, which cannot be fully utilized by encoder--decoder models, especially autoregressive ones.

Specifically, we propose to use multi-layer Transformer~\cite{attentionisallyouneed} as the non-autoregressive architecture for summarization. 
Each Transformer layer is composed of a multi-head attention sublayer and a feed-forward sublayer. 
Additionally, there is a residual connection in each sublayer, followed by layer normalization.

Let $X^{(n)} \in \mathbb{R}^{T \times d}$ be the representation at the $n$th layer, where $T$ is the number of words and $d$ is the dimension. Specially, the input layer $X^{(0)}$ is the embeddings of words.
Suppose we have $h$ attention heads. The output of the $i\text{th}$ head in the $n$th attention sublayer is
$A^{(n)}_i = \operatorname{softmax} \Big(\tfrac{Q_i K_i^{\top}}{\sqrt{d_{k}}}\Big) V_i
$, where $Q_i$, $K_i$, and $V_i$ are matrices calculated by three distinct multi-layer perceptrons (MLPs) from $X^{(n-1)}$; $d_{k}$ is the attention dimension.

Multiple attention heads are then concatenated:
\begin{equation}\nonumber
A^{(n)} = \operatorname { Concat }\big(A^{(n)}_1, \ldots,A^{(n)}_h\big) W_{O}
\end{equation}
where $W_{O} \in \mathbb{R}^{d \times d}$ is a weight matrix.

% X: T x d
% A: T x d/h

Then, we have a residual connection and layer normalization by
\begin{equation}
    \bar{A}^{(n)} = \operatorname{LayerNorm}\big(X^{(n-1)} + A^{(n)}\big)
    \label{eqn:res1}
\end{equation}
Further, an MLP sublayer processes $\bar{A}^{(n)}$, followed by residual connection and layer normalization, yielding the $n$th layer's representation
\begin{equation}
X^{(n)} = \operatorname{LayerNorm}\big(\bar{A}^{(n)} + 
\operatorname{MLP}(\bar{A}^{(n)})\big)
    \label{eqn:res2}
\end{equation} 

The last Transformer layer $X^{(N)}$ is fed to $\operatorname{softmax}$ to predict the {words of the} summary in a non-autoregressive manner, that is, the probability at the $t$th step is given by $\operatorname{softmax}(W \bm x_t^{(N)})$, where $\bm x_t^{(N)}$ is the $t$th row of the matrix $X^{(N)}$ and $W$ is the {weight matrix}.

It is emphasized that, in the vocabulary, we include a special blank token $\epsilon$, which is handled by dynamic programming during both training and inference (\S\ref{ss:DP}). This enables us to generate a shorter summary than the input with such a multi-layer Transformer.

Our model can be thought of as an encoder-only architecture, differing from a typical encoder--decoder model with cross attention~\cite{attentionisallyouneed,baziotis-etal-2019-seq,zhou-rush-2019-simple}. 
Previously, \newcite{sunon} propose a seemingly similar model to us, but put multiple end-of-sequence (EOS) tokens at the end of the generation; thus, they are unable to maintain the correspondence between input and output. Instead, we allow blank tokens scattering over the entire sentence; the residual connections in Eqns~\eqref{eqn:res1} and~\eqref{eqn:res2} can better utilize such input--output correspondence for summarization. 

\subsection{Training and Inference}\label{ss:DP}

In this section, we first introduce the Connectionist Temporal Classification (CTC) training. Then, we propose a length-control decoding approach for summary generation.

\textbf{CTC Training.}  The Connectionist Temporal Classification  \cite[CTC,][]{10.1145/1143844.1143891} algorithm allows a special blank token $\epsilon$ in the vocabulary, and uses dynamic programming to marginalize out such blank tokens{, known as \textit{latent alignment}~\cite{saharia-etal-2020-non}}. In addition, non-autoregressive generation suffers from a common problem that words may be repeated in consecutive steps~\cite{gu2017non,leedeterministic}; thus, CTC merges repeated words unless separated by $\epsilon$. For example, the sequence of tokens $a\epsilon\epsilon aabb\epsilon$ is reduced to the text $aab$, denoted by $\Gamma(a\epsilon\epsilon aabb\epsilon)=aab$.

Concretely, the predicted likelihood is marginalized over all possible fillings of $\epsilon$, i.e., all possible token sequences that are reduced to the groundtruth text:
\begin{equation}\label{eqn:marginal}
    P(\mathbf{y}|\mathbf{x})=\sum\nolimits_{\mathbf{w} : \Gamma (\mathbf w)=\mathbf y} P(\mathbf{w}|\mathbf{x})
\end{equation}
where $P(\mathbf w|\mathbf x)$ is the probability of generating a sequence of tokens $\mathbf w$. Although enumerating every candidate in $\{\mathbf w:\Gamma(\mathbf w)= \mathbf y\}$ is intractable, such marginalization fortunately can be computed by dynamic programming in an efficient way.

Let $\alpha_{s,t}=\sum_{\mathbf w_{1:s}:\Gamma(\mathbf w_{1:s})=\mathbf y_{1:t}}P(\mathbf w_{1:s}|\mathbf x)$ be the marginal probability of generating $\mathbf y_{1:t}$ up to the $s$th decoding slot. Moreover, $\alpha_{s,0}$ is defined to be the probability that $\mathbf w_{1:s}$ is all $\epsilon$, thus not having matched any word in $\mathbf y$.
The $\alpha_{s,t}$ variable can be further decomposed into two terms $\alpha_{s,t}=\alpha_{s,t}^{\epsilon}+ \alpha_{s,t}^{\neg\epsilon}$, where the first term is such probability with $\mathrm w_s=\epsilon$, and the second term $\mathrm w_s\ne\epsilon$.
Apparently, the initialization of $\alpha$ variables is
\begin{align} 
&\alpha^{\epsilon}_{1,0}= P(\mathrm {w}_1=\epsilon|\mathbf x) \\\label{eqn:init2}
&\alpha^{\neg\epsilon}_{1,1}= P(\mathrm {w}_1 = \mathrm{y}_1|\mathbf x)\\\label{eqn:init3}
&\alpha^{\epsilon}_{1,t}=0, \forall t\geq1\\\label{eqn:init4}
&\alpha^{\neg\epsilon}_{1,t}=0, \forall t>1 \text{ or } t=0
\end{align}
Eqn.~\eqref{eqn:init3} is because, at the first prediction slot, the empty token $\epsilon$ does not match any target words; Eqn.~\eqref{eqn:init4} is because the predicted non-$\epsilon$ first token must match exactly the first target word.

The recursion formula for $\alpha^{\epsilon}_{s,t}$ is
\begin{equation}  \nonumber
    \alpha^{\epsilon}_{s,t}=\alpha_{s-1,t}P(\mathrm{w}_t=\epsilon|\mathbf{x})
\end{equation}
since the newly predicted token $\epsilon$ with probability $P(\mathrm w_t=\epsilon|\mathbf x)$ does not match any target word, inheriting $\alpha_{s-1,t}$. 

The recursion formula for $\alpha^{\neg\epsilon}_{s,t}$ is
\begin{equation} \nonumber
\alpha^{\neg\epsilon}_{s,t}= \left\{\begin{array}{lc}
\left( \alpha^{\epsilon}_{s-1,t-1}  + \alpha^{\neg\epsilon}_{s-1,t} \right) P(\mathrm{w}_s=\mathrm{y}_t|\mathbf{x}), 
&\\ \quad\quad\quad\quad\quad\quad\quad\quad\quad\ \text { if } \mathrm{y}_t = \mathrm{y}_{t-1} \\
\left( \alpha_{s-1,t-1}+\alpha^{\neg \epsilon}_{s-1,t} \right) P(\mathrm{w}_s=\mathrm{y}_t|\mathbf{x}), &\\
\quad\quad\quad\quad\quad\quad\quad\quad\quad\quad  \text{ otherwise}
\end{array}\right .
\label{eqn:CTC_recursion_2}
\end{equation}
Here, $\mathrm w_s$ is not $\epsilon$, so we must have $\mathrm w_s=\mathrm y_t$, having the predicted probability $P(\mathrm w_s=\mathrm y_t|\mathbf x)$.

If $\mathrm y_t=\mathrm y_{t-1}$, then we have two sub-cases: first,
$\mathbf{w}_{1:s-1}$ is reduced to $\mathbf{y}_{1:t-1}$ with $\mathrm w_{s-1}=\epsilon$ separating two repeating words in $\mathbf y$, having probability $\alpha_{s-1,t-1}^\epsilon$; or second, 
$\mathbf{w}_{1:s-1}$ is reduced to $\mathbf{y}_{1:t}$ with  $\mathrm w_{s-1}=\mathrm y_t \ne \epsilon$, having probability $\alpha_{s-1}^{\neg\epsilon}$, which implies we are merging $\mathrm w_{s-1}$ and $\mathrm w_s$.

If $\mathrm y_t\ne \mathrm y_{t-1}$, $\mathbf{w}_{1:s-1}$ is reduced to either $\mathbf{y}_{1:t-1}$ or $\mathbf{y}_{1:t}$. In the first case, $\mathrm{w}_{s-1}$ can be either $\epsilon$ or non-$\epsilon$, given by $\alpha_{s-1,t-1}=\alpha_{s-1,t-1}^\epsilon+\alpha_{s-1,t-1}^{\neg\epsilon}$. In the second case, we must have $\mathrm{w}_{s-1} \neq \epsilon$, which has a probability of $\alpha^{\neg \epsilon}_{s-1,t}$.

Finally, $\alpha_{|\mathbf{w}|,|\mathbf{y}|}$ is the marginal probability in Eqn.~\eqref{eqn:marginal}, as it is the probability that the entire generated sequence matches the entire target text. 

The CTC maximum likelihood estimation is to maximize the marginal probability, which is equivalent to minimizing the loss $-\alpha_{|\mathbf{w}|,|\mathbf{y}|}$. Since the dynamic programming formulas are differentiable, the entire model can be trained by backpropagation in an end-to-end manner with auto-differentiation tools (such as PyTorch).

\textbf{Length-Control Inference.} 
{Controlling output length is the nature of the summarization task, for example, displaying a short news headline on a mobile device.} Moreover, \newcite{schumann-etal-2020-discrete} show that the main evaluation metric ROUGE  \cite{lin2004rouge} is sensitive to the summary length, and longer summaries tend to achieve higher ROUGE scores. Thus, it is crucial to control the summary length for fair comparison.

We propose a length-control algorithm by dynamic programming (DP), following the nature of CTC training. However, our DP is an approximate algorithm because of the dependencies introduced by removing consecutive repeated tokens. Thus, we equip our DP with a beam search mechanism.

We define $\mathscr B_{s,t}$ to be a set of top-$B$ sequences with $s$ predicted tokens that are reduced to $t$ words. $\mathscr B_{s,t}$ is constructed by three scenarios.

\begin{figure}[!t]
\centering
\includegraphics[width=.7\linewidth]{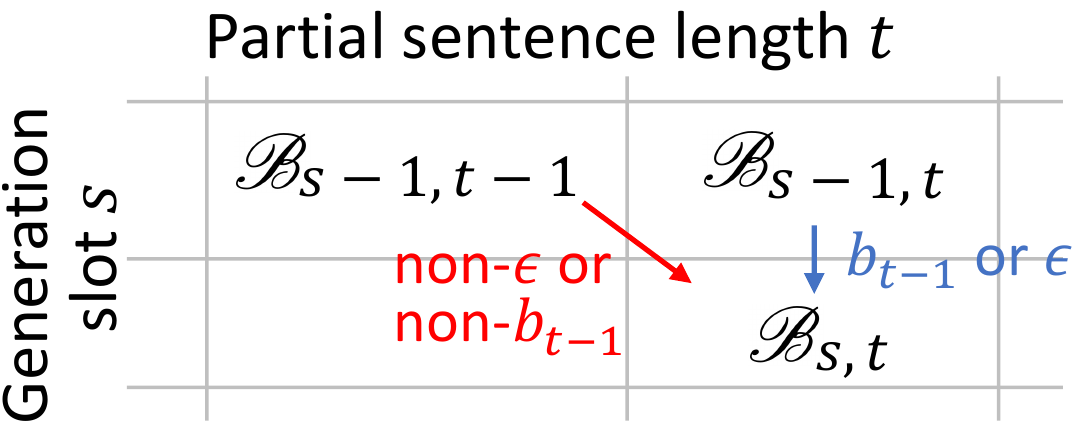}
\caption{Illustration of our length-control algorithm.}
\label{fig:length_control}
\end{figure}

First, the blank token $\epsilon$ is predicted for the $s$th generation slot, and thus the summary length $t$ remains the same, shown by the blue arrow in Figure~\ref{fig:length_control}. This yields a set of candidates
\begin{align}\label{eqn:rec1}
&\mathscr B_{s,t}^{(1)} = \big\{  \mathbf{b} \oplus \epsilon  \, : \,  \mathbf{b} \in  \mathscr B_{s-1,t}\big\}
\end{align}
where $\oplus$ refers to string/token concatenation.

Second, a repeated word is predicted for the $s$th generation slot, i.e., $\mathrm b_{s-1}$ for a subsequence $\mathbf b$ of length $s-1$. In this case, the summary length $t$ also remains the same, also shown in the blue arrow in Figure~\ref{fig:length_control}. This gives a candidate set
\begin{align}\label{eqn:rec2}
&\mathscr B_{s,t}^{(2)} = \big\{ \mathbf{b}\oplus \mathrm{b}_{s-1} \, : \, \mathbf{b} \in  \mathscr B_{s-1,t}\big\}
\end{align}

Third, a non-$\epsilon$, non-repeating word $\mathrm w_s$ is generated, increasing the summary length from $t-1$ to $t$, shown by the red arrow in Figure~\ref{fig:length_control}. This gives 
\begin{align}\nonumber
 \mathscr B_{s,t}^{(3)} = \operatorname{top}_B\big\{ \mathbf{b}\oplus \mathrm{w} \, : \, & \mathbf{b} \in  \mathscr B_{s-1,t-1},\mathrm{w}_s \neq \epsilon,\\
 &\mathrm{w}_s \neq\mathrm{b}_{s-1}\big\}
 \label{eqn:rec3}
 \end{align}
where $\operatorname{top}_B$ selects the best $B$ elements by the probability $P(\mathrm{w}_s|\mathbf x)$.

Based on the three candidates sets, we select top-$B$ sequences to keep the beam size fixed: 
\begin{align} \label{eqn:rec4}
\mathscr B_{s,t} =\operatorname{top}_B (\mathscr B_{s,t}^{(1)}\cup \mathscr B_{s,t}^{(2)} \mathscr\cup \mathscr B_{s,t}^{(3)} )
\end{align}
where the sequences are ranked by their predicted joint probabilities.

\begin{theorem}
\label{theorem:dp}
(1) If repeating tokens are not merged, then the proposed length-control algorithm with beam size $B=1$ finds the exact optimum  $\mathscr B_{S,T}$ being the most probable length-$T$ sentence given by $S$ prediction slots. (2) If we merge repeating tokens predicted by CTC-trained models, the above algorithm may not be exact.
\end{theorem}

Appendix~\ref{app:proof} presents the proof of the theorem and provides a more detailed analysis, showing that our length-control algorithm, although being approximate inference, can generate a summary of the desired length properly. Compared with truncating an overlength output, our approach is able to generate more fluent and complete sentences. Also, our length-control algorithm is different from conventional beam search, shown in Appendix~\ref{app:beam}.

\section{Experiments}
\label{sec:experiment}

\subsection{Setup}
\textbf{Datasets.} We evaluated our NAUS model on Gigaword headline generation and DUC2004 datasets.

The headline generation dataset~\cite{Rush_2015} is constructed from the Gigaword news corpus \cite{graff2003english}, where the first sentence of a news article is considered as input text and the news title is considered as the summary. The dataset contains 3.8M/198K/1951 samples for training/validation/test. Based on the analysis of the training size in Appendix~\ref{app:details}, we used 3M samples for training NAUS.

It should be emphasized that, when NAUS learns from search, we only use the input of the training corpus: we perform search~\cite{schumann-etal-2020-discrete} for each input, and train our NAUS from the search results. Therefore, we do not utilize any labeled parallel data, and our approach is unsupervised.

Moreover, we considered two settings with desired summary lengths of 8 and 10, following~\newcite{schumann-etal-2020-discrete}. Our NAUS is trained from respective search results.

The DUC2004 dataset \cite{duc2004} is designed for testing only with 500 samples, where we also take the first sentence of an article as the input text. Our NAUS is transferred from the above headline generation corpus. Based on the length of DUC2004 summaries, we trained NAUS from search results with 13 words, also following \newcite{schumann-etal-2020-discrete} for fair comparison.

\textbf{Evaluation Metrics.} We evaluated the quality of predicted summaries by ROUGE scores \footnote{https://github.com/tagucci/pythonrouge} ~\cite{lin2004rouge}, which are the most widely used metrics in previous work~\cite{wang-lee-2018-learning, baziotis-etal-2019-seq, zhou-rush-2019-simple}.
Specifically, ROUGE-$n$ evaluates $n$-gram overlap between a predicted summary and its reference summary; ROUGE-L, instead, measures the longest common sequence between the predicted and reference summaries. 

Different ROUGE variants are adopted in previous work, depending on the dataset. We followed the standard evaluation scripts and evaluated headline generation by ROUGE F1~\cite{wang-lee-2018-learning,baziotis-etal-2019-seq,schumann-etal-2020-discrete} and DUC2004 by Truncate ROUGE Recall~\cite{dorr-etal-2003-hedge,west-etal-2019-bottlesum}. 

In addition to summary quality, we also evaluated the inference efficiency of different methods, as it is important for the deployment of deep learning models in real-time applications. 
We report the average inference time in seconds for each data sample, and compare the speedup with \newcite{schumann-etal-2020-discrete}'s search approach, which achieves (previous) state-of-the-art ROUGE scores.
Our experiments were conducted on an i9-9940X CPU and an RTX6000 graphic card.
 Appendix~\ref{app:details} presents additional implementation details.
\begin{table*}[!t]
\centering
\resizebox{\textwidth}{!}{%
\begin{tabular}{|c|c|c|c|r|rrrr|rr|}
\hline
\multirow{2}{*}{Group} &
  \multirow{2}{*}{\#} &
    \multicolumn{2}{c|}{\multirow{2}{*}{Approach}} &
  \multirow{2}{*}{Len} &
  \multicolumn{4}{c|}{ROUGE F1} &
  \multirow{2}{*}{Inf.Time} &
  \multirow{2}{*}{Speedup} \\ \cline{6-9}
 &
   &
   \multicolumn{2}{c|}{}
   &
   &
  R-1 &
  R-2 &
  R-L &
  $\Delta$R &
   &
   \\ \hline\hline

\multirow{5}{*}{\begin{tabular}[c]{@{}c@{}}A\\ (desired\\ length 8)\end{tabular}} &
  1 &
  \multirow{1}{*}{Baseline} &
  $\text{Lead (8 words)}^\dagger$ &
  7.9 &
  21.39 &
  7.42 &
  20.03 &
  -11.12 &
  -- &
  -- 
  \\\cline{2-11} 
 &
  2 &
   \multirow{2}{*}{Search} &
  $\text{\newcite{schumann-etal-2020-discrete}}^\dagger$ &
  7.9 &
  26.32 &
  9.63 &
  24.19 &
  0.18 &
  -- &
  -- \\ 
 &
  3 &
    &
  Our replication &
  7.9 &
  26.17 &
  \textbf{9.69} &
  24.10 &
  0 &
  6.846 &
  1x \\ \cline{2-11} 
 &
  4 &
  \multicolumn{1}{c|}{\multirow{3}{*}{\begin{tabular}[c]{@{}c@{}}Learn from \\ search\end{tabular}}} &
  \newcite{sunon} &
  7.7 &
  26.88 &
  9.37 &
  24.54 &
  0.83 &
  0.017 &
  403x \\
  & 5 & &
  NAUS (truncate) &
  7.8 &
  27.27 &
  9.49 &
  24.96 &
  1.76 &
  \textbf{0.005} &
  \textbf{1369x} \\
 &
  6 &
   &
  NAUS (length control) &
  7.8 &
  \textbf{27.94} &
  9.24 &
  \textbf{25.51} &
  \textbf{2.73} &
  0.041 &
  167x \\ \hline\hline
\multirow{8}{*}{\begin{tabular}[c]{@{}c@{}}B\\ (desired \\ length 10)\end{tabular}} &
  7 &
  \multirow{3}{*}{Baseline} &
  $\text{Lead (10 words)}^\dagger$ &
  9.8 &
  23.03 &
  7.95 &
  21.29 &
  -10.2 &
  -- &
  -- \\
 &
  8 &
   &
  $\text{\newcite{wang-lee-2018-learning}}^\dagger$ &
  10.8 &
  27.29 &
  10.01 &
  24.59 &
  -0.58 &
  -- &
  -- \\
 &
  9 &
   &
  $\text{\newcite{zhou-rush-2019-simple}}^\dagger$ &
  9.3 &
  26.48 &
  10.05 &
  24.41 &
  -1.53 &
  -- &
  -- \\ 
  \cline{2-11} 
 &
  10 &
  \multirow{2}{*}{Search} &
  $\text{\newcite{schumann-etal-2020-discrete}}^\dagger$ &
  9.8 &
  27.52 &
  \textbf{10.27} &
  24.91 &
  0.23 &
  -- &
  -- \\ 
 &
  11 &
   &
  Our replication &
  9.8 &
  27.35 &
  10.25 &
  24.87 &
  0 &
  9.217 &
  1x \\ \cline{2-11} 
 &
  12 &
  \multicolumn{1}{c|}{\multirow{3}{*}{\begin{tabular}[c]{@{}c@{}}Learn from \\ search\end{tabular}}} &
  \newcite{sunon} &
  9.4 &
  27.86 &
  9.88 &
  25.51 &
  0.78 &
  0.020 &
  461x\\
  & 13& &
  NAUS (truncate) &
  9.8 &
  28.24 &
  10.04 &
  25.40 &
  1.21 &
  \textbf{0.005} &
  \textbf{1843x} \\
 &
  14 &
   &
  NAUS (length control) &
  9.8 &
  \textbf{28.55} &
  9.97 &
  \textbf{25.78} &
  \textbf{1.83} &
  0.044&
  210x \\\hline
\end{tabular}
}
\caption{Results on the Gigaword headline generation test set. 
\textbf{Len:} Average length of predicted summaries. \textbf{R-1, R-2, R-L:} ROUGE-1, ROUGE-2, ROUGE-L. \textbf{$\Delta$R:} The difference of total ROUGE (sum of R-1, R-2, and R-L) in comparison with the (previous) state-of-the-art search method under replication.
\textbf{Inf.Time:} Average inference time in seconds for one sample on an i9-9940X CPU and a RTX6000 GPU.
\textbf{Speedup:} Relative to \newcite{schumann-etal-2020-discrete}.
$^\dagger$Results quoted from previous papers; others are given by our experiments.
} 

\label{table:giga_performance}
\end{table*}

\subsection{Results and Analyses}

\textbf{Main Results.} Table \ref{table:giga_performance} presents the performance of our model and baselines on the Gigaword headline test set. For a fair comparison, we categorize all approaches by average summary lengths of \textasciitilde8 and \textasciitilde10 into Groups A and B, respectively. 

The Lead baseline extracts the first several words of the input sentence. Despite its simplicity, the Lead approach is a strong summarization baseline adopted in most previous work \cite{fevry-phang-2018-unsupervised,baziotis-etal-2019-seq}. 

\begin{table}[!t]
\centering
\resizebox{0.5\textwidth}{!}{%
\begin{tabular}{|c|rrrr|rr|}
\hline
\multirow{2}{*}{Model} &
  \multicolumn{4}{c|}{ROUGE Recall} & \multirow{2}{*}{\!\!\!\!Time} &
  \multicolumn{1}{l|}{\multirow{2}{*}{\!\!\!\!Speedup}\!\!} \\ \cline{2-5}
 &
  \multicolumn{1}{c}{R-1} &
  \multicolumn{1}{c}{R-2} &
  \multicolumn{1}{c}{R-L} &
  \multicolumn{1}{c|}{$\Delta$R} & &
   \\ \hline\hline
$\text{Lead (75 characters)}^\dagger$                                   & 22.50          & 6.49          & 19.72          & -8.34     & --    & --    \\
$\text{\newcite{zajic2004bbn}}^\dagger$                & 25.12          & 6.46          & 20.12          & -5.35     & --    & --    \\
$\text{\newcite{baziotis-etal-2019-seq}}^\dagger$      & 22.13          & 6.18          & 19.30          & -9.44     & --    & --    \\
$\text{\newcite{west-etal-2019-bottlesum}}^\dagger$    & 22.85          & 5.71          & 19.87          & -8.62     & --    & --    \\
\hline
$\!\!\text{\newcite{schumann-etal-2020-discrete}}^\dagger$\!\!\!\! & 26.04          & \textbf{8.06} & 22.90          & -0.05         & -- & --    \\ 
\!\!Our replication\!\!\!   &26.14   &8.03  &22.88   & 0 &\!\!\!\!12.314 & 1x \\
\hline
\newcite{sunon}    & 26.25          & 7.66          & 22.83          & -0.31     & 0.022    & 559x    \\
NAUS (truncate)                                        & 26.52          & 7.88          & 22.91          & 0.26      & \textbf{0.005}  & \textbf{2463x} \\
NAUS (length control)                                  & \textbf{26.71} & 7.68          & \textbf{23.06} & \textbf{0.40} & 0.048 & 257x \\\hline
\end{tabular}%
}
\caption{Results on the DUC2004 dataset. $^\dagger$Quoted from previous papers.}
\label{table:duc2004_performance}
\end{table}

\newcite{wang-lee-2018-learning} utilize cycle consistency~\cite{miao2016language} for unsupervised summarization; {the performance is relatively low, because the cycle consistency loss cannot ensure the generated text is a valid summary.}
\newcite{zhou-rush-2019-simple} perform beam search towards a step-by-step decomposable score of fluency and contextual matching.
Both are unable to explicitly control the summary length: in a fair comparison of length 10 (Group~B, Table~\ref{table:giga_performance}), their performance is worse than the (previous) state-of-the-art approach \cite{schumann-etal-2020-discrete},\footnote{\newcite{schumann-etal-2020-discrete} present a few variants that use additional datasets for training language models (in an unsupervised way).
In our study, we focus on the setting without data augmentation, i.e., the language model is trained on non-parallel the Gigawords corpus.} 
which performs edit-based local search.

Our NAUS approach follows \newcite{schumann-etal-2020-discrete}, but trains a non-autoregressive model from search results. 
We consider two settings for controlling the summary length: truncating longer summaries and decoding with our proposed length-control algorithm.
Both of our variants outperform \newcite{schumann-etal-2020-discrete} by 1.21--2.73 in terms of the total ROUGE score (Rows~5--6 \& 13--14, Table~\ref{table:giga_performance}). As mentioned, \newcite{schumann-etal-2020-discrete} only extract original words with order preserved,  yielding noisy sentences. Our NAUS, as a student, learns from the search-based teacher model and is able to smooth out its noise. This is a compelling result, as our student model outperforms its teacher.

Regarding inference efficiency, our NAUS method with truncating is more than 1300 times faster than \newcite{schumann-etal-2020-discrete}, because we do not need iterative search.
Even with dynamic programming and beam search for length control, NAUS is still over 100 times faster. 
This shows our NAUS is extremely efficient in inference, which is important for real-time applications.

Although the efficiency of \newcite{wang-lee-2018-learning} and \newcite{zhou-rush-2019-simple} is not available, we still expect our approach to be a few times faster (despite our higher ROUGE scores) because their models are autoregressive. By contrast, our NAUS is non-autoregressive, meaning that it predicts all words simultaneously. We will provide a controlled comparison between autoregressive and non-autoregressive models in Table~\ref{table:giga_ablation_result}.

Table~\ref{table:duc2004_performance} shows the results on the DUC2004
dataset. 
The cycle-consistency approach \cite{baziotis-etal-2019-seq,west-etal-2019-bottlesum} does not perform well on this dataset, outperformed by an early rule-based syntax tree trimming approach \cite{zajic2004bbn} and the state-of-the-art edit-based search~\cite{schumann-etal-2020-discrete}. 

The performance of our NAUS model is consistent with Table \ref{table:giga_performance}, outperforming all previous methods in terms of the total ROUGE score, and being 100--1000 times faster than the search approach \cite{schumann-etal-2020-discrete}.

In general, the proposed NAUS not only achieves state-of-the-art ROUGE scores for unsupervised summarization, but also is more efficient when deployed. Results are consistent on both datasets, demonstrating the generality of our NAUS.

\textbf{In-Depth Analyses.}
We conduct in-depth analyses on the proposed NAUS model in Table~\ref{table:giga_ablation_result}. Due to the limit of time and space, we chose the Gigaword headline generation as our testbed. All the autoregressive (AR) and non-autoregressive (NAR) variants learn from the search output of our replication (Rows~2~\&~11), where we achieve very close results to those reported in \newcite{schumann-etal-2020-discrete}.

\begin{table}[!t]
\centering
\resizebox{0.5\textwidth}{!}{%
\begin{tabular}{|c|ccrrrrr|}
\hline 
\multicolumn{1}{|c|}{\multirow{2}{*}{\#}} & \multicolumn{2}{|c|}{\multirow{2}{*}{Approach}} &
  \multicolumn{4}{c|}{ROUGE Recall} &
  \multicolumn{1}{l|}{\multirow{2}{*}{\!\!\!Speedup}\!\!\!} \\ \cline{4-7}
\multicolumn{1}{|c|}{} &
\multicolumn{2}{|c|}{} &
  \multicolumn{1}{c}{R-1} &
  \multicolumn{1}{c}{R-2} &
  \multicolumn{1}{c}{R-L} &
  \multicolumn{1}{c|}{$\Delta$R} &
  \multicolumn{1}{l|}{} \\ \hline\hline\multicolumn{8}{|c|}{Group A (desired length 8)} \\ \hline
  1& 
\multicolumn{1}{|c|}{\multirow{2}{*}{Search}} &
  \multicolumn{1}{c|}{\!\!\text{\citeauthor{schumann-etal-2020-discrete}}\!\!} &
  26.32 &
  9.63 &
  24.19 &
  \multicolumn{1}{r|}{0.18} &
  -- \\
  2 &
\multicolumn{1}{|c|}{} &
  \multicolumn{1}{c|}{\!\!Our replication\!\!} &
  26.17 &
  \textbf{9.69} &
  24.10 &
  \multicolumn{1}{r|}{0} &
  1x \\ \hline
  3 & 
\multicolumn{1}{|c|}{AR} &
  \multicolumn{1}{c|}{Transformer (T)} &
  26.65 &
  9.51 &
  24.67 &
  \multicolumn{1}{r|}{0.87} &
  58x \\ \hline
  4&
\multicolumn{1}{|c|}{\multirow{3}{*}{\begin{tabular}[c]{@{}c@{}}NAR\\ enc-dec\end{tabular}}} &
  \multicolumn{1}{c|}{Vanilla} &
  24.87 &
  8.33 &
  22.74 &
  \multicolumn{1}{r|}{\!\!\!\!-4.02} &
  571x \\
  5 &
\multicolumn{1}{|c|}{} &
  \multicolumn{1}{c|}{CTC (T)} &
  27.30 &
  9.20 &
  24.96 &
  \multicolumn{1}{r|}{1.5} &
  571x \\
  6 &
\multicolumn{1}{|c|}{} &
  \multicolumn{1}{c|}{CTC (LC)} &
  27.76 &
  9.13 &
  25.33 &
  \multicolumn{1}{r|}{2.26} &
  149x \\ \hline
    7 &
\multicolumn{1}{|c|}{\multirow{3}{*}{\begin{tabular}[c]{@{}c@{}}NAR\\ \!\!\!enc-only\!\!\!\end{tabular}}} &
  \multicolumn{1}{c|}{\newcite{sunon}}
   &
  26.88 &
  9.37 &
  24.54 &
  0.83 &
  \multicolumn{1}{|c|}{403x}\\
  8& &
  \multicolumn{1}{|c|}{Our NAUS (T)} &
  27.27 &
  9.49 &
  24.96 &
  \multicolumn{1}{r|}{1.76} &
  \!\!\textbf{1396x} \\
  9 &
\multicolumn{1}{|c|}{} &
  \multicolumn{1}{c|}{Our NAUS (LC)} &
  \textbf{27.94} &
  9.24 &
  \textbf{25.51} &
  \multicolumn{1}{r|}{2.73} &
  167x \\ \hline\hline

\multicolumn{8}{|c|}{Group B (desired length 10)} \\ \hline
10 &
\multicolumn{1}{|c|}{\multirow{2}{*}{Search}} &
  \multicolumn{1}{c|}{\!\!\!\text{\citeauthor{schumann-etal-2020-discrete}}\!\!\!} &
  27.52 &
  \textbf{10.27} &
  24.91 &
  \multicolumn{1}{r|}{0.23} &
  -- \\
  \!\!11\!\! &
\multicolumn{1}{|c|}{} &
  \multicolumn{1}{c|}{\!\!\!Our replication\!\!\!} &
  27.35 &
  10.25 &
  24.87 &
  \multicolumn{1}{r|}{0} &
  1x \\ \hline
  \!\!12\!\! &
\multicolumn{1}{|c|}{AR} &
  \multicolumn{1}{c|}{Transformer (T)} &
  27.06 &
  9.63 &
  24.55 &
  \multicolumn{1}{r|}{\!\!\!\!-1.23} &
  66x \\ \hline
  \!\!13\!\! &
\multicolumn{1}{|c|}{\multirow{3}{*}{\begin{tabular}[c]{@{}c@{}}NAR\\ enc-dec\end{tabular}}} &
  \multicolumn{1}{c|}{Vanilla} &
  25.77 &
  8.69 &
  23.52 &
  \multicolumn{1}{r|}{\!\!\!\!-4.49} &
  709x \\
  \!\!14\!\! &
\multicolumn{1}{|c|}{} &
  \multicolumn{1}{c|}{CTC (T)} &
  28.14 &
  10.07 &
  25.37 &
  \multicolumn{1}{r|}{1.11} &
  709x \\
  \!\!15\!\! &
\multicolumn{1}{|c|}{} &
  \multicolumn{1}{c|}{CTC (LC)} &
  28.45 &
  9.81 &
  25.63 &
  \multicolumn{1}{r|}{1.42} &
  192x \\ \hline
  \!\!16\!\! &
\multicolumn{1}{|c|}{\multirow{3}{*}{\begin{tabular}[c]{@{}c@{}}NAR\\ \!\!\!enc-only\!\!\!\end{tabular}}} &
  \multicolumn{1}{|c|}{\newcite{sunon}} &
  27.86 &
  9.88 &
  25.51 &
  0.78 &
  \multicolumn{1}{|c|}{461x} \\
  17& &
  \multicolumn{1}{|c|}{Our NAUS (T)} &
  28.24 &
  10.04 &
  25.40 &
  \multicolumn{1}{r|}{1.21} &
  \!\!\textbf{1843x} \\
  \!\!18\!\! &
\multicolumn{1}{|c|}{} &
  \multicolumn{1}{c|}{Our NAUS (LC)} &
  \textbf{28.55} &
  9.97 &
  \textbf{25.78} &
  \multicolumn{1}{r|}{\textbf{1.83}} &
  210x \\ \hline
\end{tabular}%
}
\caption{Model analysis on headline generation. \textbf{AR:} Autoregressive models. \textbf{NAR enc-dec:} Non-autoregressive encoder--decoder. \textbf{NAR enc-only:} Non-autoregressive encoder-only. \textbf{T:} Truncating. \textbf{LC:} Length control.  All AR and NAR models use the Transformer architecture.}
\label{table:giga_ablation_result}
\end{table}

We first tried vanilla encoder--decoder NAR Transformer \cite[Rows~4~\&~13,][]{gu2017non}, where we set the number of decoding slots as the desired summary length; thus, {the blank token} and the length-control algorithm are not needed. As seen, a vanilla NAR model does not perform well, and CTC largely outperforms vanilla NAR in both groups (Rows 5--6 \& 14--15). Such results are highly consistent with the translation literature~\cite{saharia-etal-2020-non,imputer,gu-kong-2021-fully,qian2020glancing,huang2021non}.

The proposed encoder-only NAUS model outperforms encoder--decoder ones in both groups in terms of the total ROUGE score, when the summary length is controlled by either truncating or length-control decoding (Rows 8--9 \& 17--18). 
Profoundly, our non-autoregressive NAUS is even better than the autoregressive Transformer (Rows~3 \&~12). 
We also experimented with previous non-autoregressive work for supervised summarization~\cite{sunon}\footnote{To the best of our knowledge, the other two non-autoregressive supervised summarization models are \newcite{yang-etal-2021-pos} and \newcite{pmlr-v139-qi21a}. Their code and pretrained models are not available, making replication difficult.} in our learning-from-search setting. Although their approach appears to be encoder-only, it adds end-of-sequence (EOS) tokens at the end of the generation, and thus is unable to utilize the input--output correspondence. Their performance is higher than vanilla NAR models, but lower than ours. By contrast, NAUS is able to capture such correspondence with the residual connections, i.e., Eqns.~\eqref{eqn:res1} and~\eqref{eqn:res2}, in its encoder-only architecture.

Generally, the efficiency of encoder-only NAR\footnote{The standard minimal encoder--decoder NAR model has 6 layers for the encoder and another 6 layers for the decoder~\cite{attentionisallyouneed}. Our NAUS only has a 6-layer encoder. Our pilot study shows that more layers do not further improve performance in our encoder-only architecture.} (without length-control decoding) is \textasciitilde2 times faster than encoder--decoder NAR and \textasciitilde20 times faster than the AR Transformer.

Further, our length-control decoding improves the total ROUGE score, compared with truncating, for both encoder--decoder CTC and encoder-only NAUS models (Rows~6, 9, 15, \& 18), although its dynamic programming is slower. 
Nevertheless, our non-autoregressive NAUS with length control is \textasciitilde200 times faster than search and \textasciitilde3 times faster than the AR Transformer.

\textbf{Additional Results.} We present additional results in our appendices:

\ref{app:beam}. Analysis of Beam Search

\ref{app:case}. Case Study

\ref{app:human}. Human Evaluation

\ref{app:transfer}. Length-Transfer Summarization

\section{Related Work}

Summarization systems can be generally categorized into two paradigms: extractive and abstractive. Extractive systems extract certain sentences and clauses from input, for example, based on salient features~\cite{zhou-rush-2019-simple} or feature construction~\cite{he2012document}. Abstraction systems generate new utterances as the summary, e.g., by sequence-to-sequence models trained in a supervised way~\cite{zhang2020pegasus,liurefsum}. 

Recently, unsupervised abstractive summarization is attracting increasing attention. \newcite{yang-etal-2020-ted} propose to use the Lead baseline (first several sentences) as the pseudo-groundtruth. However, such an approach only works with well-structured articles (such as CNN/DailyMail). \newcite{wang-lee-2018-learning} and \newcite{baziotis-etal-2019-seq} use cycle consistency for unsupervised summarization. \newcite{zhou-rush-2019-simple} propose a step-by-step decomposable scoring function and perform beam search for summary generation. \newcite{schumann-etal-2020-discrete} propose an edit-based local search approach, which allows a more comprehensive scoring function and outperforms cycle consistency and beam search. 

Our paper follows \newcite{schumann-etal-2020-discrete} but trains a machine learning model to improve efficiency and smooth out search noise. Previously, \newcite{NEURIPS2020_7a677bb4} fine-tune a GPT-2 model based on search results for unsupervised paraphrasing; \newcite{jolly2021search} adopt the search-and-learning framework to improve the semantic coverage for few-shot data-to-text generation. We extend previous work in a non-trivial way by designing a non-autoregressive generator and further proposing a length-control decoding algorithm.

{The importance of controlling the output length is recently realized in the summarization community. \newcite{baziotis-etal-2019-seq} and \newcite{sunon} adopt soft penalty to encourage shorter sentences; \newcite{yang-etal-2021-pos} and \newcite{pmlr-v139-qi21a} control the summary length through POS tag and EOS predictions. 
None of these studies can control the length explicitly.  
\newcite{song-etal-2021-new} is able to precisely control the length by progressively filling a pre-determined number of decoding slots, analogous to the vanilla NAR model in our non-autoregressive setting. 
}

Non-autoregressive generation is originally proposed for machine translation~{\cite{gu2017non, guo2020fine, saharia-etal-2020-non}}{, which is later extended to other text generation tasks. \newcite{wiseman-etal-2018-learning} address the table-to-text generation task, and model output segments by a hidden semi-Markov model \cite{ostendorf1996hmm}, simultaneously generating tokens for all segments.}
\newcite{jia2021flexible} apply non-autoregressive models to extractive document-level summarization. \newcite{sunon} stack a non-autoregressive BERT model with a conditional random field (CRF) for abstractive summarization; since the summary is shorter than the input text, their approach puts multiple end-to-sequence (EOS) tokens at the end of the sentence, and thus is unable to utilize the strong input--output correspondence in the summarization task. \newcite{yang-etal-2021-pos} apply auxiliary part-of-speech (POS) loss and \newcite{pmlr-v139-qi21a} explore pretraining strategies for encoder--decoder non-autoregressive summarization. All these studies concern supervised summarization, while our paper focuses on unsupervised summarization. We adopt CTC training in our encoder-only architecture, allowing blank tokens to better align input and output words, which is more appropriate for summarization.

\section{Conclusion}

In this work, we propose a non-autoregressive unsupervised summarization model (NAUS), where we further propose a length-control decoding algorithm based on dynamic programming. 
Experiments show that NAUS not only archives state-of-the-art unsupervised performance on Gigaword headline generation and DUC2004 datasets, but also is much more efficient than search methods and autoregressive models. Appendices present additional analyses and length-transfer experiments.

\textbf{Limitation and Future Work.} Our paper focuses on unsupervised summarization due to the importance of low-data applications. One limitation is that we have not obtained rigorous empirical results for supervised summarization, where the developed model may also work. This is because previous supervised summarization studies lack explicit categorization of summary lengths~\cite{yang-etal-2020-ted,pmlr-v139-qi21a}, making comparisons unfair and problematic \cite{schumann-etal-2020-discrete}. Such an observation is also evidenced by \newcite{sunon}, where the same model may differ by a few ROUGE points when generating summaries of different lengths. Nevertheless, we have compared with \newcite{sunon} in our setting and show the superiority of the NAUS under fair comparison. We plan to explore supervised summarization in future work after we establish a rigorous experimental setup, which is beyond the scope of this paper.

\section{Acknowledgments}
We thank Raphael Schumann for providing valuable suggestions on the work. We also thank the Action Editor and reviewers for their comments during ACL Rolling Review. The research is supported in part by the Natural Sciences and Engineering Research Council of Canada (NSERC) under grant
No.~RGPIN2020-04465, the Amii Fellow Program, the Canada CIFAR AI Chair Program, a UAHJIC project, a donation from DeepMind, and Compute Canada (www.computecanada.ca).

\bibliography{custom}
\bibliographystyle{acl_natbib}

\appendix

\section{Proof of Theorem \ref{theorem:dp}} 
\label{app:proof}

\noindent\textbf{Theorem 1.}
\textit{(1) If repeating tokens are not merged, then the proposed length-control algorithm with beam size $B=1$ finds the exact optimum  $\mathscr B_{S,T}$ being the most probable length-$T$ sentence given by $S$ prediction slots. (2) If we merge repeating tokens predicted by CTC-trained models, the above algorithm may not be exact.}

\medskip
\begin{proof}[Proof.]

[Part (1)] This part concerns a variant of our decoding algorithm, which only removes the blank token $\epsilon$ but does not merge consecutive repeated tokens to a single word, i.e., Eqn.~\eqref{eqn:rec2} is removed. We denote this by $\Gamma'$, for example, $\Gamma'(a\epsilon\epsilon aa bb\epsilon)=aaabb$, as opposed to $\Gamma(a\epsilon\epsilon aa bb\epsilon)=aab$ in our algorithm. We now show that, based on $\Gamma'$, our dynamic programming algorithm in \S\ref{ss:DP} with beam size $B=1$ is an exact inference algorithm.

We define $\beta_{s,t}=\max_{\mathbf b: |\mathbf b|=s,|\Gamma'(\mathbf b)|=t} P(\mathbf b|\mathbf x)$, where $|\cdot|$ denotes the length of a sequence. In other words, $\beta_{s,t}$ is the maximum probability of $s$ tokens that are reduced to $t$ words.

According to the definition, we have
\begin{align}
&\beta_{1,0}= P\left(\mathrm{w}_1=\epsilon|\mathbf x\right) \label{eqn:l_control_init1}\\
&\beta_{1,1}= \max\nolimits_{\mathrm{w}_1 \neq \epsilon} P\left(\mathrm{w}_1 |\mathbf x\right) \label{eqn:l_control_init2}\\ \label{eqn:beta_init3}
&\beta_{s,t}=0 \text{\quad for $s> t$}
\end{align}
In \eqref{eqn:l_control_init1}, $\beta_{1,0}$ refers to the probability of one token that is reduced to zero words, in which case the first predicted token can only be the blank token $\epsilon$,  corresponding to Eqn.~\eqref{eqn:rec1} with $s=1$ and $t=0$. Likewise, $\beta_{1,1}$ is the maximum probability of one token that is reduced to one word. Thus, it is the probability of the most probable non-$\epsilon$ token, corresponding to Eqn.~(\ref{eqn:rec3}) with $s=1$ and $t=0$. Eqn.~\eqref{eqn:beta_init3} asserts that fewer tokens cannot be reduced to more words; it is used for mathematical derivations, but need not to be explicitly implemented in our algorithm in \S\ref{ss:DP}.

The recursion variable $\beta_{s,t}$ is computed by
\begin{equation}
\begin{aligned}
\beta_{s,t}= 
\max \Big\{&\beta_{s-1,t} \cdot  P(\mathrm{w}_s=\epsilon |\mathbf{x}) ,\\ &\beta_{s-1,t-1} \cdot\max\nolimits_{\mathrm{w}_s \neq \epsilon} P(\mathrm{w}_s |\mathbf x) \Big\} \label{eqn:beta}
\end{aligned}
\end{equation}
In other words, the variable $\beta_{s,t}$ can inherit $\beta_{s-1,t}$ with a predicted blank token $\epsilon$, corresponding to Eqn.~\eqref{eqn:rec1}; or it can inherit $\beta_{s-1,t-1}$ with a predicted non-$\epsilon$ token, corresponding to Eqn.~\eqref{eqn:rec3}. Specially, if $t=0$, then the second term has $\beta_{s-1,-1}$ undefined, and thus is ignored in the $\max$ operation.

We need the $\max$ operator to take the higher probability in the two cases, 
since $\beta_{s,t}$ is the maximum probability of $s$ tokens being reduced to $t$ words. This corresponds to Eqn.~\eqref{eqn:rec4} with beam size $B=1$.

To sum up, our inductive calculation guarantees that $\beta_{S,T}$ is the exact maximum probability of $\max_{\mathbf b: |\mathbf b|=S,|\Gamma'(\mathbf b)|=T} P(\mathbf b|\mathbf x)$ for the desired length $T$ with $S$ generation slots; our algorithm (if not merging repeating tokens) gives  the corresponding $\mathscr B_{S,T}$ as $\operatorname{argmax} P(\mathbf b|\mathbf x)$ under the same constraints, concluding the proof of Part (1).

[Part (2)] CTC training merges consecutive repeated tokens to a single word, unless separated by the blank token $\epsilon$~\cite{10.1145/1143844.1143891}. Since our model is trained by CTC, we should adopt this rule in inference as well. We show in this part that our algorithm, with beam size $B=1$, may not yield the exact optimum with an example in Table~\ref{table:greedy_illustration}.

\begin{table}[!t] 
\centering
\resizebox{.5\linewidth}{!}{
\begin{tabular}{|c|c|c|}
\hline Word & $P(\mathrm{w_1}|\mathbf{x})$ & $P(\mathrm{w_2}|\mathbf{x})$ \\ \hline\hline
I & 0.39 & 0.1   \\ \hline
like & 0.4   & 0.9  \\ \hline
coding & 0.1  & 0   \\ \hline
$\epsilon$ & 0.11  & 0  \\ \hline
\end{tabular}}
\caption{An example of predicted probabilities of two generation slots, where we have a vocabulary of three words and a blank token $\epsilon$.}\label{table:greedy_illustration}
\end{table}

We consider generating a sentence of two words from the two prediction slots, i.e., $S=T=2$. Apparently, the optimal sequence is ``I like'' with probability $0.39\cdot 0.9=0.351$. However, the algorithm would predict $\mathscr B_{1,1}=\{\text{``like''}\}$ because ``like'' is the most probably token in the first slot. Then, our algorithm will give $\mathscr B_{2,2}=\{\text{``like I''}\}$, because it has to select a non-repeating token based on $\Gamma$, yielding a non-optimal solution.

\end{proof}

It is noted that, if we do not merge repeating tokens as in $\Gamma'$, our algorithm will give the exact optimum ``like like'' in the above example. This shows that merging consecutive repeated tokens requires the decoding algorithm to correct early predictions, and thus, our dynamic programming becomes an approximate inference. Nevertheless, our algorithm is able to generate a sequence of the desired length properly; its approximation happens only when the algorithm compares more repetitions with fewer $\epsilon$s versus more $\epsilon$s with fewer repetitions. Such approximation is further alleviated by beam search in our dynamic programming. Therefore, the proposed length-control algorithm is better than truncating a longer sentence; especially, our approach generates more fluent and complete sentences.

\section{Implementation Details} \label{app:details}

Our NAUS had a Transformer encoder as the basic structure, generally following the settings in \newcite{attentionisallyouneed}: 6 encoder layers, each having 8 attention heads. The dimension was 512 for attention and 2048 for feed-forward modules. 

Our training used a batch size of 4K tokens, with a maximum of 200K updates. We used Adam with $\beta=(0.9, 0.98)$.
In general, the learning rate warmed up to 5e-4 in the ﬁrst 10K steps, and then decayed to 1e-9 with the inverse square-root schedule, except that we find the maximum learning rate of 1e-4 worked better for headline generation with the summary length of 8. We set the $\ell_2$ weight decay to 0.01. Our length-control decoding algorithm had a beam size of 6. More details can be found in our repository (Footnote 1). 

\begin{figure}[!t]
\centering
\includegraphics[width=0.6\linewidth]{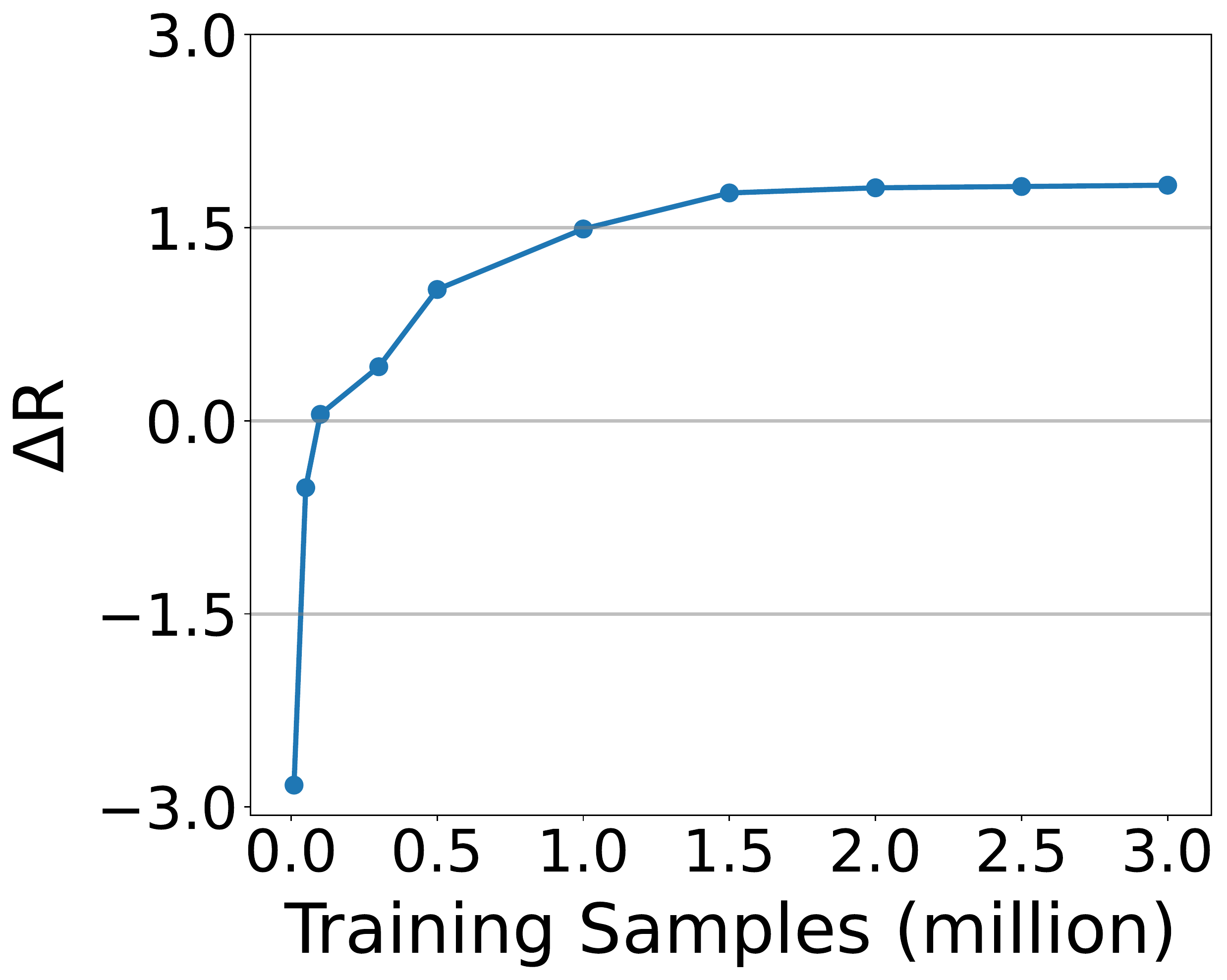}
\caption{Performance versus the number of training samples in the setting of Group B, Table~\ref{table:giga_performance}. Notice that NAUS is trained by pseudo-groundtruth given by unsupervised edit-based search~\cite{schumann-etal-2020-discrete}. Thus, our approach is indeed unsupervised.}
\label{fig:performance_diff_data}
\end{figure}

Our NAUS training is based on \newcite{schumann-etal-2020-discrete}'s prediction on the input of the Gigaword headline generation training set. We show performance against the number of training samples in Figure~\ref{fig:performance_diff_data}. As seen, NAUS outperforms its search teacher even with a small set of 0.1 million samples. The performance saturates as the number of samples increases. Based on this analysis, we used 3 million samples from the 3.8 million Gigaword training set to train our NAUS models. 

Each reported number in Tables~\ref{table:giga_performance}--\ref{table:giga_ablation_result} were averaged over 10 independent runs, whereas the results in Table~\ref{tab:giga_length_transfer} (Appendix~\ref{app:transfer}) were based on a single run due to the limited time. 

\section{Analysis of Beam Search}\label{app:beam}

As mentioned, our length-control decoding algorithm involves beam search within its dynamic programming, because the algorithm does not find the exact optimum when it merges repeating words. We analyze the effect of the beam size in our length-control algorithm.

In addition, we compare our approach with CTC beam search~\cite{10.1145/1143844.1143891}.\footnote{Our implementation of CTC beam search is based on \url{https://github.com/parlance/ctcdecode}}
Typically, a CTC-trained non-autoregressive model can be decoded either greedily or by beam search. The greedy decoding finds the most probable token at each step, i.e., $\mathrm w_i^*=\operatorname{argmax}_{\mathrm{w}_i}P( \mathrm w_i|\mathbf x)$, and reduces the tokens to a sentence by $\Gamma(\mathrm w_1,\cdots, \mathrm w_T)$, where $T$ is the number of decoding steps. The CTC beam search algorithm searches for the most likely sentence by marginalizing all token sequences that are reduced to $\mathbf y$, i.e., $\argmax_{\mathbf y}\sum_{\mathbf w: \Gamma(\mathbf w)=\mathbf y}P(\mathbf w|\mathbf x)$.

\begin{figure}[!t]
\centering
\includegraphics[width=1\linewidth]{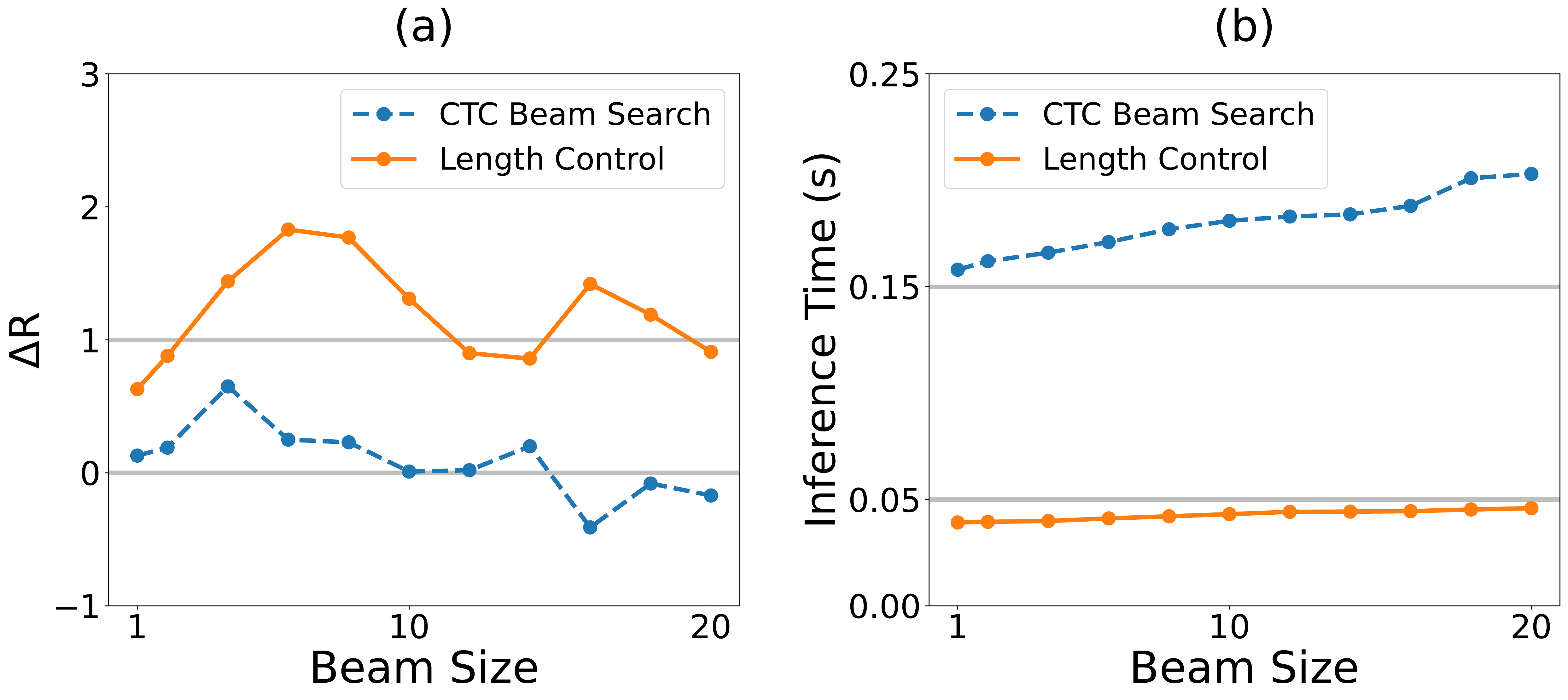}
\caption{Comparing our length-control NAUS and the truncated CTC beam search on the Gigaward headline generation test set.}
\label{fig:beam_search}
\end{figure}

We show results in Figure~\ref{fig:beam_search}, where we chose 10-word Gigaword headline generation as the testbed with our NAUS model (Group B, Table~\ref{table:giga_performance}). Notice that CTC beam search does not control the output length, and for fair comparison, we truncated its generated summaries. This also shows that our novel decoding approach and CTC beam search are distinct algorithms.

As seen in Figure~\ref{fig:beam_search}a, the beam search does play a role in our length-control algorithm. When the beam enlarges from 1 to 6, the performance (orange solid line) increases by 1.2 points in $\Delta$R, the difference of total ROUGE in comparison with \newcite{schumann-etal-2020-discrete} under our replication (Row 10, Table~\ref{table:giga_performance}). However, further increasing the beam size does not yield additional performance gain. This is consistent with previous literature in autoregressive generation~\cite{meister2020if}, which also suggests a beam size of 5--7 is the best in their applications. In terms of the efficiency (Figure~\ref{fig:beam_search}b), a larger beam size monotonically increases the inference time. However, the overhead of beam search is relatively small in our dynamic programming, and thus we chose a beam size of 6 in our experiments.

Our length-control algorithm significantly outperforms CTC beam search (dashed blue lines) in terms of both $\Delta$R and efficiency. Especially, CTC beam search is three times slower, and degrades more significantly than our length-control decoding when the beam size increases.

\section{Case Study} \label{app:case}
We show in Table~\ref{tab:case_study} example summaries generated by our NAUS with truncating and length-control decoding, as well as the previous state-of-the-art method \cite{schumann-etal-2020-discrete}. We observe that NAUS without length control generates slightly longer summaries, and if truncated, the output may be incomplete; by contrast, our length-control algorithm can generate a fluent and complete sentence of the desired length by dynamic programming.
Compared with \newcite{schumann-etal-2020-discrete}, our NAUS (length control) generates a more informative summary that includes the main clause (\textit{united nations condemned}), which also appears in the reference summary.

\section{Human Evaluation}
\label{app:human}
\begin{table}[!t]
\centering
\resizebox{0.5\textwidth}{!}{%
\begin{tabular}{|c|c|rrr|r|}
\hline
 &
  Decoding &
  \multicolumn{1}{c}{Wins} &
  \multicolumn{1}{c}{Ties} &
  \multicolumn{1}{c|}{Loses} &
  \multicolumn{1}{l|}{$p$-val} \\ \hline
\multirow{2}{*}{\!\!Overall quality\!\!} & Truncate       & 18.6{7}\%          & 40.6{7}\% & 40.6{7}\% & \multirow{2}{*}{\!\!0.0004\!\!} \\
                                 & \!\!Length control\!\! & \textbf{40.6{7}\%} & 40.6{7}\% & \textbf{18.6{7}}\% &                        \\ \hline
\multirow{2}{*}{\begin{tabular}[c]{@{}c@{}}\!\!\!Completeness\!\!\! \\ \& fluency\end{tabular}} &
  Truncate &
  24.6{7}\% &
  26.6{7}\% &
  48.6{7}\% &
  \multirow{2}{*}{\!\!0.0005\!\!} \\
                                 & \!\!Length control\!\! & \textbf{48.6{7}\%} & 26.6{7}\% & \textbf{24.6{7}}\% &                        \\ \hline
\end{tabular}%
}
\caption{Human evaluation comparing truncating and length control for our NAUS model on 50 samples in the Gigaword headline generation task. The results are statistically significant, where the $p$-value is given by a one-sided binomial test.}
\label{tab:human_evaluation_result}
\end{table}

We conducted human evaluation with a focus on truncating and length-control decodings. This is because truncating may generate incomplete sentences, which cannot be adequately evaluated by automatic metrics as their ROUGE scores are close.

Specifically, we invited three human annotators to compare the two decoding algorithms for NAUS on 50 randomly selected samples, in the setting of Group B, Table~\ref{table:giga_performance} (Gigaword headline generation with a target length of 10). The annotation was conducted in a pairwise manner in terms of overall quality and fluency/completeness; average results (wins/loses/ties) are shown in Table~\ref{table:greedy_illustration}. It should be mentioned that our annotation was strictly blind: the samples of two systems were presented in random order and annotators did not know which system generated a sample.

As seen, our length-control decoding algorithm largely outperforms the truncating approach in terms of both the overall quality and fluency/completeness. The results are statistically significant ($p$-values $<0.01$) in a one-sided binomial test.  
This verifies that length-control decoding is important for summarization, as truncating yields incomplete sentences, which are inadequately reflected by ROUGE scores.

\begin{table}[!t]
\centering
\resizebox{0.45\textwidth}{!}{%
\begin{tabular}{|lllll|}
\hline
\multicolumn{5}{|l|}{\begin{tabular}[c]{@{}l@{}}\textbf{Input:} the united nations condemned saturday an attack on \\ russian embassy employees in baghdad that claimed the life \\ of one russian and resulted in the kidnapping of four others\end{tabular}} \\
\multicolumn{5}{|l|}{\begin{tabular}[c]{@{}l@{}}\textbf{Reference:} un condemns murder of russians in iraq with \\ annan comment\end{tabular}}                              \\
\multicolumn{5}{|l|}{\begin{tabular}[c]{@{}l@{}}\textbf{\newcite{schumann-etal-2020-discrete}:}  attack on russian embassy in \\ baghdad claimed one in four\end{tabular}} \\
\multicolumn{5}{|l|}{\begin{tabular}[c]{@{}l@{}}\textbf{NAUS (truncate):} an attack on russian embassy employees \\ in baghdad claimed in {\color{gray} kidnapping of four others}\end{tabular}}                    \\
\multicolumn{5}{|l|}{\begin{tabular}[c]{@{}l@{}}\textbf{NAUS (length control):} united nations condemned attack \\ on russian embassy employees in baghdad\end{tabular}}   \\ \hline
\end{tabular}%
}
\caption{Example summaries for Gigaword headline generation. The gray words are truncated for fair comparison.}
\label{tab:case_study}
\end{table}

\begin{table*}[!t]
\centering
\resizebox{1\textwidth}{!}{%
\begin{tabular}{|c|c|cc|r|rrrr|rr|}
\hline
\multirow{2}{*}{Group} &
  \multirow{2}{*}{\#} &
  \multicolumn{2}{c|}{\multirow{2}{*}{Approach}} &
  \multirow{2}{*}{Len} &
  \multicolumn{4}{c|}{ROUGE F1} &
  \multicolumn{1}{l}{\multirow{2}{*}{Inf.Time}} &
  \multicolumn{1}{l|}{\multirow{2}{*}{Speedup}} \\ \cline{6-9}
 &
   &
  \multicolumn{2}{c|}{} &
   &
  R-1 &
  R-2 &
  R-L &
  $\Delta$R &
  \multicolumn{1}{l}{} &
  \multicolumn{1}{l|}{} \\ 
 \hline \hline
\multirow{7}{*}{\begin{tabular}[c]{@{}c@{}}Group A\\ (desired length 8)\end{tabular}} &
  1 &
  \multicolumn{1}{c|}{Baseline} &
  $\text{Lead (8 words)}^\dagger$ &
  7.9 &
  21.39 &
  7.42 &
  20.03 &
  -11.12 &
  -- &
  -- \\ \cline{2-11} 
 &
  2 &
  \multicolumn{1}{c|}{\multirow{2}{*}{Search}} &
  $\text{\newcite{schumann-etal-2020-discrete}}^\dagger$ &
  7.9 &
  26.32 &
  9.63 &
  24.19 &
  0.18 &
  -- &
  -- \\
 &
  3 &
  \multicolumn{1}{c|}{} &
  Our replication &
  7.9 &
  26.17 &
  \textbf{9.69} &
  24.10 &
  0 &
  6.846 &
  1x \\ \cline{2-11} 
  
     &
  4 &
   \multicolumn{1}{c|}{\multirow{5}{*}{\begin{tabular}[c]{@{}c@{}}Learn from \\ search\end{tabular}}}&
  \newcite{sunon}$_{8\rightarrow 8}$ &
  7.7 &
  26.88 &
  9.37 &
  24.54 &
  0.83 &
  0.017 &
  403x \\
     &
  5 &
  \multicolumn{1}{c|}{} &
  \newcite{sunon}$_{10\rightarrow 8}$ &
  8.4 &
  25.71 &
  8.94 &
  23.65 &
  -1.84 &
  0.018 &
  380x \\
  
 &
  6 &
  \multicolumn{1}{c|}{} &
  NAUS (truncate) &
  7.8 &
  27.27 &
  9.49 &
  24.96 &
  1.76 &
  \textbf{0.005} &
  \textbf{1369x} \\
 &
  7 &
  \multicolumn{1}{c|}{} &
  NAUS$_{8\rightarrow 8}$ &
  7.8 &
  \textbf{27.94} &
  9.24 &
  \textbf{25.50} &
  \textbf{2.73} &
  \multirow{2}{*}{0.041} &
  \multirow{2}{*}{167x} \\
 &
  8 &
  \multicolumn{1}{c|}{} &
  NAUS$_{10\rightarrow 8}$ &
  7.9 &
  27.12 &
  9.08 &
  24.86 &
  1.10 &
   &
   \\ \hline \hline
  \multirow{8}{*}{\begin{tabular}[c]{@{}c@{}}\\Group B\\ (desired  length 10)\end{tabular}} &
  9 &
  \multicolumn{1}{c|}{\multirow{3}{*}{Baseline}} &
  $\text{Lead (10 words)}^\dagger$ &
  9.8 &
  23.03 &
  7.95 &
  21.29 &
  -10.2 &
  -- &
  -- \\
 &
  10 & \multicolumn{1}{c|}{}
   &
  $\text{\newcite{wang-lee-2018-learning}}^\dagger$ &
  10.8 &
  27.29 &
  10.01 &
  24.59 &
  -0.58 &
  -- &
  -- \\
 &
 11 & \multicolumn{1}{c|}{}
   &
  $\text{\newcite{zhou-rush-2019-simple}}^\dagger$ &
  9.3 &
  26.48 &
  10.05 &
  24.41 &
  -1.53 &
  -- &
  -- \\ 
  \cline{2-11} 
 &
  12 &
  \multicolumn{1}{c|}{\multirow{2}{*}{Search}} &
  $\text{\newcite{schumann-etal-2020-discrete}}^\dagger$ &
  9.8 &
  27.52 &
  \textbf{10.27} &
  24.91 &
  0.23 &
  -- &
  -- \\ 
 &
  13 & \multicolumn{1}{c|}{}
   &
  Our replication &
  9.8 &
  27.35 &
  10.25 &
  24.87 &
  0 &
  9.217 &
  1x \\ \cline{2-11} 
     &
  14 & \multicolumn{1}{c|}{\multirow{5}{*}{\begin{tabular}[c]{@{}c@{}}Learn from \\ search\end{tabular}}}
   &
  \newcite{sunon}$_{8\rightarrow 10}$ &
  -- &
  -- &
  -- &
  -- &
  -- &
  -- &
  -- \\
     &
  15 & \multicolumn{1}{c|}{}
   &
  \newcite{sunon}$_{10\rightarrow 10}$ &
  9.4 &
  27.86 &
  9.88 &
  25.51 &
  0.78 &
  0.020 &
  461x \\
 &
  16 &
   \multicolumn{1}{c|}{}
   &
  NAUS (truncate) &
  9.8 &
  28.24 &
  10.04 &
  25.40 &
  1.21 &
  \textbf{0.005} &
  \textbf{1843x} \\
 &
  17 & \multicolumn{1}{c|}{}
   &
  NAUS$_{8\rightarrow 10}$ &
  9.9 &
  28.32 &
  9.58 &
  25.46 &
  0.89 &
 \multirow{2}{*}{0.044} &
  \multirow{2}{*}{210x} \\
  & 18 & \multicolumn{1}{c|}{}
   &
  NAUS$_{10\rightarrow 10}$ &
  9.8 &
  \textbf{28.55} &
  9.97 &
  \textbf{25.78} &
  \textbf{1.83} &
  &
   \\

  \hline \hline
  \multirow{6}{*}{\begin{tabular}[c]{@{}c@{}}\\Group C\\ (desired length \\ 50\% of the input)\end{tabular}} &
  19 &
  \multicolumn{1}{c|}{\multirow{3}{*}{Baseline}} &
  $\text{Lead (50\% words)}^\dagger$ &
  14.6 &
  24.97 &
  8.65 &
  22.43 &
  -4.58 &
  -- &
  -- \\
 &
  20 &
  \multicolumn{1}{c|}{} &
  $\text{\newcite{fevry-phang-2018-unsupervised}}^\dagger$ &
  14.8 &
  23.16 &
  5.93 &
  20.11 &
  -11.43 &
  -- &
  -- \\
 &
  21 &
  \multicolumn{1}{c|}{} &
  $\text{\newcite{baziotis-etal-2019-seq}}^\dagger$ &
  15.1 &
  24.70 &
  7.97 &
  22.41 &
  -5.55 &
  -- &
  -- \\ \cline{2-11} 
 &
  22 &
  \multicolumn{1}{c|}{\multirow{2}{*}{Search}} &
  $\text{\newcite{schumann-etal-2020-discrete}}^\dagger$ &
  14.9 &
  27.05 &
  9.75 &
  23.89 &
  0.06 &
  -- &
  -- \\
 &
  23 &
  \multicolumn{1}{c|}{} &
  Our replication &
  14.9 &
  27.03 &
  9.81 &
  23.79 &
  0 &
  17.462 &
  1x \\ \cline{2-11} 
   &
 24 &
   \multicolumn{1}{c|}{\multirow{4}{*}{\begin{tabular}[c]{@{}c@{}}Learn from \\ search\end{tabular}}}
 & 
  \newcite{sunon}$_{8\rightarrow 50\%}$ &
  -- &
  -- &
  -- &
  -- &
  -- &
  -- &
  -- \\
   &
 25 &
   \multicolumn{1}{c|}{}
 & 
  \newcite{sunon}$_{10\rightarrow 50\%}$ &
  -- &
  -- &
  -- &
  -- &
  --&
  -- &
  -- \\
 &
 26 &
   \multicolumn{1}{c|}{}
 & 
  NAUS$_{8\rightarrow 50\%}$ &
  14.9 &
  28.39 &
  9.78 &
  24.94 &
  2.48 &
  \multirow{2}{*}{\textbf{0.052}} &
  \multirow{2}{*}{\textbf{336x}} \\
  &
  27 & \multicolumn{1}{c|}{}
     &
  NAUS$_{10\rightarrow 50\%}$ &
  14.9 &
  \textbf{28.53} &
  \textbf{9.88} &
  \textbf{25.10} &
  \textbf{2.88} &
   &
   \\
  \hline
\end{tabular}%
}
\caption{Analysis of length-transfer summary generation. A subscript ${i\rightarrow j}$ (or ${j\%})$ refers to a model trained with $i$ words and tested for $j$ (or $j\%$) words. \textbf{Len:} Average length of predicted summaries. \textbf{R-1, R-2, R-L:} ROUGE-1, ROUGE-2, ROUGE-L. \textbf{$\Delta$R:} The difference of total ROUGE (sum of R-1, R-2, and R-L) in comparison with the (previous) state-of-the-art model~\cite{schumann-etal-2020-discrete} under replication.
\textbf{Inf.Time:} Average inference time in seconds for one sample on an i9-9940X CPU and a RTX6000 GPU.
\textbf{Speedup:} Relative to \newcite{schumann-etal-2020-discrete}.
$^\dagger$Results quoted from previous papers; others are given by our experiments. \newcite{sunon}'s approach has a soft length penalty to encourage short output, but cannot generate longer summaries than trained.}
\label{tab:giga_length_transfer}
\end{table*}

\section{Length-Transfer Summary Generation}\label{app:transfer}

In the main paper, we present results where our NAUS is trained on search outputs~\cite{schumann-etal-2020-discrete} that have the same length as the inference target.
This follows the common assumption in machine learning that training and test samples are independently identically distributed.

In this appendix, we show the performance of length-transfer summary generation, where the prediction has a different length from that of training. We denote such a model by NAUS${}_{i\rightarrow j}$, referring to training with $i$ words and testing for $j$ words. 

As seen in Groups A \& B in Table~\ref{tab:giga_length_transfer}, NAUS with length transfer is slightly worse than NAUS trained on the correct length, which is understandable. Nevertheless, length-transfer decoding still outperforms the search teacher and other baselines. 

Moreover, we consider the third setting in \newcite{schumann-etal-2020-discrete}, where the target length is 50\% of the input. Since it takes time to obtain pseudo-groundtruths given by the edit-based search, we would directly transfer already trained NAUS models to this setting by our length-control decoding. Results are shown in Group C, Table~\ref{tab:giga_length_transfer}.
We observe NASU${{}_{10\rightarrow 50\%}}$ is better than NASU${{}_{8\rightarrow 50\%}}$, which makes much sense because the latter has a larger gap during transfer. Remarkably, both NASU${{}_{8\rightarrow 50\%}}$ and NASU${{}_{10\rightarrow 50\%}}$ outperform \newcite{schumann-etal-2020-discrete} and other baselines, achieving new state-of-the-art unsupervised performance on this setting as well. 

We further compare with \newcite{sunon}, who use a length penalty to encourage short summaries. However, their length control works in the statistical sense but may fail for individual samples. Moreover, such a soft length penalty cannot generate longer summaries than trained. Even in the setting of $10\rightarrow 8$, their generates summaries are slightly longer than required, while the performance degrades much more considerably than NAUS.

These results show that our novel length-control decoding algorithm is not only effective when generating summaries of similar length to the training targets, but also generalizes well to different desired summary lengths without re-training. In general, our NAUS is an effective and efficient unsupervised summarization system with the ability of explicit length control.

\end{document}